\documentclass[nonatbib]{article}

\usepackage[final]{neurips_2024}

\usepackage[utf8]{inputenc} 
\usepackage[T1]{fontenc}    
\usepackage{hyperref}       
\usepackage{url}            
\usepackage{booktabs}       
\usepackage{amsfonts}       
\usepackage{nicefrac}       
\usepackage{microtype}      
\usepackage{xcolor}         
\usepackage{amsmath}
\usepackage[pdftex]{graphicx}

\title{From Text to Treatment Effects: A Meta-Learning Approach to Handling Text-Based Confounding}

\author{%
  Henri Arno \\
  Ghent University - imec\\
  \texttt{henri.arno@ugent.be} \\
   \And
   Paloma Rabaey \\
  Ghent University - imec\\
  \texttt{paloma.rabaey@ugent.be} \\
   \And
   Thomas Demeester \\
  Ghent University - imec\\
  \texttt{thomas.demeester@ugent.be}
}

\begin{document}

\maketitle


\begin{abstract}
One of the central goals of causal machine learning is the accurate estimation of heterogeneous treatment effects from observational data. In recent years, meta-learning has emerged as a flexible, model-agnostic paradigm for estimating \textit{conditional} average treatment effects (CATE) using any supervised model. This paper examines the performance of meta-learners when the confounding variables are expressed in text. Through synthetic data experiments, we show that learners using pre-trained text representations of confounders, in addition to tabular background variables, achieve improved CATE estimates compared to those relying solely on the tabular variables, particularly when sufficient data is available. However, due to the entangled nature of the text embeddings, these models do not fully match the performance of meta-learners with perfect confounder knowledge. These findings highlight both the potential and the limitations of pre-trained text representations for causal inference and open up interesting avenues for future research.
\end{abstract}

\section{Introduction}

Treatment effects can vary substantially across different subgroups within a population. Accurately estimating these heterogeneous effects can guide decision-making in a wide range of critical areas such as personalised medicine and public policy. For instance, doctors need to identify which patients are most likely to benefit from specific treatments. 
Likewise, governments must determine who would gain most from programs such as subsidised job training. 
Although randomised controlled trials (RCTs) are the gold standard for estimating these effects, ethical and practical constraints often limit their feasibility. Recent advances in machine learning have enabled data-driven estimation of heterogeneous treatment effects from observational data \cite{alaa2024conformal, gausian_processes, curth2021nonparametric, hill2011bayesian, kennedy2023towards, kunzel2019metalearners, nie2021quasi, shalit3, shi2019adapting, wager2018estimation,yoon2018ganite}.


Since individual-level treatment effects are unobservable—known as \textit{the fundamental problem of causal inference}—estimating treatment effects differs from traditional supervised learning due to the absence of a direct prediction target. One approach proposed in literature is meta-learning, which addresses this challenge by decomposing treatment effect estimation into separate sub-problems that can each be tackled with standard supervised models \cite{kunzel2019metalearners}. Alternatively, various machine learning techniques have been adapted for estimating heterogeneous treatment effects, including Gaussian processes \cite{gausian_processes}, random forests \cite{wager2018estimation} and GANs \cite{yoon2018ganite}.

Recent advances in meta-learning have broadened its applicability to a wider range of problems. For instance, new methods enable meta-learners to provide predictive intervals to account for uncertainty around the point-estimates of heterogeneous treatment effects \cite{alaa2024conformal, jonkers2024conformal} or to estimate these effects over time \cite{frauen2024}. Following this line of work, our paper explores how meta-learners perform when the confounding variables are expressed in text. This is particularly relevant given that many real world applications involve unstructured data. For instance, in personalised medicine, diagnostic information in electronic health records is often recorded in clinical notes written by physicians. Similarly, in public policy, career characteristics of individuals are usually accessible only through survey or social media data. This leads to our central research question: \textit{How do meta-learners perform with pre-trained text representations of confounders, and how does this compare to settings where confounders are either ignored or perfectly known?}

\section{Background and related work}

\paragraph{Problem definition:}
We position our work in the Rubin-Neyman framework on causality \cite{rubinneyman}, where heterogeneous treatment effects can be formalised as \textit{conditional} average treatment effects (CATE). Consider the observed data $\{ \left(X_i, T_i, Y_i^{obs}\right)\}_{i=1}^D$, where $X_i$ represents the covariates for unit~$i$, potentially including confounders, $T_i$ is the treatment indicator with $P(T_i=1|X_i)=\pi(X_i)$ (i.e., the propensity score) and $Y_i^{obs}$ is the observed outcome. In this framework, each unit $i$ also has two \textit{potential} outcomes, $Y_i(0)$ and $Y_i(1)$, representing the outcomes we would observe under no treatment and treatment, respectively. The CATE is the expected difference between potential outcomes, conditioned on covariates $X$, and formally defined as:
\begin{equation}
    \tau(X) = \mathbb{E}[Y(1) - Y(0) | X]
\end{equation}
Under the conventional assumptions of \textit{consistency} (i.e., $Y^{obs}=Y(1)T + Y(0)(1-T)$), \textit{positivity} (i.e., $\pi(X) \in (c, 1-c)$ for $0<c<1$) and \textit{unconfoudedness} (i.e., $Y(0), Y(1) \perp \!\!\! \perp T | X$), the CATE is identifiable, meaning that it can be estimated from observed data, and can be expressed as:
\begin{equation}
    \tau(X) = \mathbb{E}[Y|T=1, X] - \mathbb{E}[Y|T=0, X]
\end{equation}

\paragraph{Considered meta-learners:}

As discussed, meta-learning decomposes CATE estimation into sub-problems that can each be addressed with standard supervised learning methods. This typically involves estimating \textit{nuisance parameters} $\hat{\eta}(X)$, which are then transformed into a \textit{pseudo-outcome} $\tilde{\tau}(X)$. Pseudo-outcomes aim to provide a noisy but potentially unbiased approximation of the CATE and can be used as targets for a second-stage regressor \cite{fisherinverse}.

In this study, we consider four established meta-learners: the T-learner \cite{kunzel2019metalearners}, the RA-learner \cite{curth2021nonparametric}, the DR-learner \cite{kennedy2023towards}, and the R-learner \cite{nie2021quasi}. These meta-learners rely on a common set of nuisance parameters $\hat{\eta}(X) = \{\hat{\mu}_0(X), \hat{\mu}_1(X), \hat{\mu}(X), \hat{\pi}(X)\}$, where $\hat{\mu}_t(X)$ is an estimate of the conditional outcome given treatment $T=t$ and covariates $X$ (i.e., $\mathbb{E}[Y|T=t, X]$), $\hat{\mu}(X)$ is an estimate of the overall conditional outcome given covariates $X$ (i.e., $\mathbb{E}[Y|X]$), and $\hat{\pi}(X)$ is an estimate of the propensity score. 
The pseudo-outcomes for the RA-, DR- and R-learner are detailed in Appendix~\ref{appendix:POs}, along with a brief overview of their theoretical background. In contrast, the T-learner directly estimates the CATE as the difference between $\hat{\mu}_1(X)$ and $\hat{\mu}_0(X)$. 
For a detailed discussion on these learners and their connections, see \cite{morzywolek2023general}.

\paragraph{Causal inference with learned representations:}

Using pre-trained text representations of confounders for CATE estimation is, to the best of our knowledge, a novel approach. However, an extensive body of literature exists on learning representations with neural networks, particularly from structured data, for causal inference. For instance, Shalit et al. \cite{shalit3} introduced a model that learns a shared representation from the covariates, with two separate regression heads to predict outcomes under treatment and no treatment respectively. They also introduced a regularization term to balance the representations during training, such that the induced distributions of the two treatment groups become similar. Building on this, Shi et al. \cite{shi2019adapting} proposed a neural network that similarly learns a shared representation from the covariates, but with a third head to also predict the propensity score. 
Several other models have been proposed to learn representations for causal inference. Curth et al. \cite{curth2021inductive} explored a range of these, specifically evaluating their effectiveness for CATE estimation within the meta-learner framework. In related work, Melnychuk et al. \cite{melnychuk2023bounds} studied \textit{representation-induced confounder bias} which arises when the learned representations lose information about the observed confounders (e.g., due to dimensionality reduction) from a theoretical perspective.

Closely related to our work, Veitch et al. \cite{veitch2020adapting} develop a method for estimating causal effects by adjusting for confounding features of text, such as subject and writing quality. Their approach adapts language models to learn text representations that are predictive of both treatment and outcome. While their work is highly relevant, our paper differs in several key aspects. We focus on confounders expressed in text rather than text features themselves, and we specifically integrate text representations within the meta-learning framework for CATE estimation, rather than focusing on average treatment effects.

\section{Data}
\vspace{-0.25cm}

\paragraph{Current benchmarking practices:}
As noted, estimating heterogeneous treatment effects is challenging because these effects are unobserved, which also complicates evaluation. To address this, the literature has typically relied on simulated data where ground-truth treatment effects are known. This can be achieved with completely synthetic data \cite{fisherinverse}, semi-synthetic data (where only the potential outcomes are simulated) \cite{shalit3}, or by fitting a generative model on real data \cite{athey}.

A widely used benchmark in the machine learning community for evaluating CATE estimators is the semi-synthetic IHDP benchmark \cite{hill2011bayesian}. This benchmark was constructed from real data of a randomised study, from which a non-random subsample of the treated units was removed to simulate confounding. The potential outcomes are generated using a relatively simple model. However, the benchmark has several limitations (see \cite{curth2021really} for a discussion), including the unknown treatment assignment mechanism and the generative process that is not representative for the real world and systematically favours certain algorithms over others.  Additionally, it is not suited for our purpose, as it lacks unstructured data. Due to these constraints, we have chosen to 
use the synthetic SynSUM benchmark instead (see below), which does include text by design, has a fully known generative process, and is more realistic, having been developed in close collaboration with a domain expert.

\vspace{-0.2cm}
\paragraph{The SynSUM benchmark:}
SynSUM \cite{SynSUM} is a dataset of 10.000 synthetic medical patient records, containing both structured tabular variables and unstructured clinical text notes describing a fictional patient encounter in the domain of respiratory diseases in primary care. 
The tabular variables include two possible diagnoses (\texttt{pneumonia} and \texttt{common cold}), four underlying respiratory conditions (\texttt{asthma}, \texttt{smoking}, \texttt{COPD} and \texttt{hay fever}), five symptoms (\texttt{dyspnea}, \texttt{cough}, \texttt{pain}, \texttt{fever} and \texttt{nasal}) and three non-clinical variables (\texttt{policy}, \texttt{self-employed} and \texttt{season}). In the fictional setting, a treatment of \texttt{antibiotics} is prescribed based on the severity of the symptoms. The outcome is \texttt{days at home}, describing how many days the patient ends up staying home as a result of their symptoms and the prescribed treatment. The five symptoms act as confounders between \texttt{antibiotics} and \texttt{days at home}. All other variables (like diagnoses and underlying conditions) either directly or indirectly influence the occurrence of the symptoms, but do not act as direct confounders between treatment and outcome.

The tabular variables (including treatment and outcome) were sampled from a Bayesian network, where both the structure and the conditional probability distributions were defined by an expert through domain knowledge. Afterwards, GPT4-o was prompted to generate a clinical note to accompany the tabular patient record. The prompt contained information on the symptoms experienced by the patient, as well as any underlying health conditions the patient may have, but no information on the diagnosis, treatment or outcome variables. 
For further details on prompting and the full directed acyclic graph relating all tabular variables, we refer the reader to \cite{SynSUM}. An example of a SynSUM entry can be found in Appendix~~\ref{app:SynSum}.

In our experiments, the five symptoms play the role of text-based confounders, while \texttt{antibiotics} is the treatment and \texttt{days at home} is the outcome. The other tabular variables are passed to the models as additional background information. The diagnosis variables (\texttt{pneumonia} and \texttt{common cold}) and \texttt{policy} (reflecting a clinician’s inclination to prescribe antibiotics) are excluded and assumed unknown to simulate a realistic setting, as these are typically unavailable at the time of prescribing treatment. Importantly, these excluded variables are no direct confounders in the dataset. As the data-generating process is fully known, the ground-truth CATE for each sample is available. The potential outcomes, $Y(0)$ and $Y(1)$, were generated using two separate Poisson regression models, where the mean number of days at home, $\mathbb{E}[Y(t)|X]$, parameterises each model and is a function of the symptoms \cite{SynSUM}.

\vspace{-0.15cm}
\section{Experimental results}

\begin{figure}[]
    \centering
    \includegraphics[width=\textwidth]{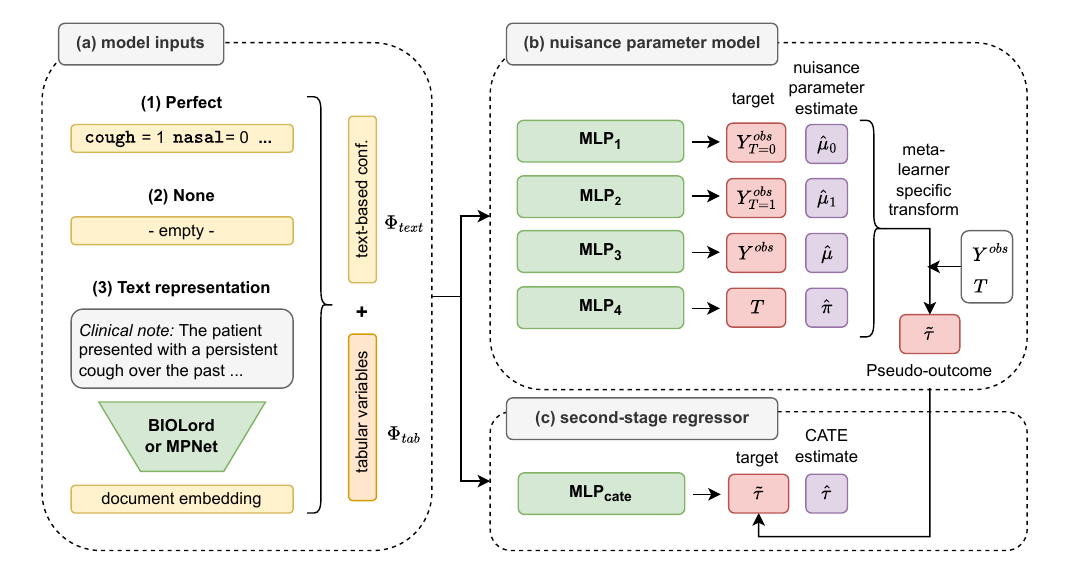} 
    \caption{\label{fig:models} Overview of the experimental setup across three panels: Panel (a) presents the different types of representations $\Phi_{text}$ of the text-based confounders, concatenated with the tabular variables $\Phi_{tab}$ to form the model inputs. Panel (b) illustrates the architecture used to estimate the nuisance parameters, which are then transformed—along with the observed outcomes and treatment indicators—into the pseudo-outcomes for each learner (except the T-learner). Panel (c) depicts the second-stage regressor, which uses these pseudo-outcomes as targets to estimate the CATE.}
\end{figure}

\vspace{-0.18cm}

\subsection{Impact of the text-based confounders on the performance of the meta-learners}
\vspace{-0.18cm}

\paragraph{Objective:} 
The aim of our initial experiments is to evaluate how the considered meta-learners perform (1) with perfect knowledge of the text-based confounders and (2) with no access to them, having to rely solely on the tabular background variables to estimate the CATE. By varying the amount of training data, we seek to determine how much data is required for each learner before a significant performance gap emerges between these two settings. This will clarify the conditions under which information on the confounders substantially improves CATE estimates and when pre-trained text representations of confounders may potentially be beneficial.

\paragraph{Experimental setup and model architecture:}
In the first setting, where we assume perfect knowledge of the text-based confounders, the CATE is estimated using both the tabular variables ($\Phi_{tab}$) and the true confounder values ($\Phi_{text}$). In the second setting, where the confounders are unknown, the model relies solely on the tabular variables ($\Phi_{tab}$). This setup is illustrated in panel (a) of Figure~\ref{fig:models}. We vary the size of the training set across 300 - 1,000 - 3,000 and 9,000 samples, while keeping a fixed test set of 1,000 samples. In both settings, these inputs are first used to estimate the nuisance parameters using four separate multi-layer perceptrons (MLPs), each with a single hidden layer of 10 units and a ReLU activation. Each MLP is trained with a target specific to the nuisance parameter being estimated. The estimated nuisance parameters are then used to construct pseudo-outcomes for each learner (except the T-learner), as shown in panel (b) of Figure~\ref{fig:models}. Finally, a second-stage regressor uses these pseudo-outcomes as targets to estimate the CATE. Like the nuisance parameter models, this regressor is an MLP with a single hidden layer of 10 units and a ReLU activation, as depicted in panel (c) of Figure~\ref{fig:models}. Consistent with Curth et al. \cite{curth2021nonparametric}, cross-fitting is not applied as the focus is on empirical performance rather than theoretical guarantees. Detailed training procedures for both the nuisance parameter models and the second-stage regressors are provided in Appendix~\ref{app:training_details}.

\paragraph{Performance evaluation and results:} For each meta-learner and training set size, the entire training process—which includes the nuisance parameter model and second-stage regressor—is repeated five times with different random seeds to account for variations in weight initialisation and data sampling. We evaluate the CATE estimates on the test set by comparing them to the ground-truth CATE using the root mean squared error, commonly referred to as \textit{precision in estimation of heterogeneous effects} (PEHE) in the context of CATE estimation \cite{hill2011bayesian}.

We present our results in Figure~\ref{fig:baselines} (Appendix~\ref{appendix:results}) which shows the performance of each meta-learner in both settings—with perfect vs. no knowledge of the text-based confounders—across various training set sizes. A clear trend emerges: as the amount of training data increases, the performance gap between the two settings widens. When the training set is small, 
models with perfect knowledge of the confounders perform similarly to those relying solely on the tabular background variables.\footnote{Except for the T-learner trained with 300 samples. In this case, the model with perfect knowledge of the confounders performs significantly better compared to the model with no knowledge of the confounders.} However,~as the training set grows, the models with perfect knowledge of the confounders continue to improve steadily, whereas the models with no access to the confounders show little or no improvement. This pattern consistently holds across all meta-learners. Additionally, we observe that the variability of the PEHE decreases for larger training sets. Notably, the DR-learner and R-learner exhibit higher variability compared to the T-learner and RA-learner, likely due to their reliance on propensity score estimates in their pseudo-outcomes (see e.g. \cite{fisherinverse}). 

These preliminary results indicate that the information on the text-based confounders only significantly improves CATE estimates when enough training data is available for the models to effectively leverage this. This provides a foundation for further experiments to investigate the potential of pre-trained text representations to improve meta-learner performance.

\subsection{Text-based confounders as pre-trained text representations}

\paragraph{Objective:} Building on the insights from our initial experiments, we now explore the potential of pre-trained text representations of confounders in improving the quality of CATE estimates. Specifically, we examine how pre-trained embeddings—both from generic and domain-specific encoders—affect the performance of meta-learners when the true confounder values are unknown. 
When sufficient data is available, and confounder information significantly improves CATE estimates, we aim to determine how the learners perform with these representations. Conversely, when little data is available, and the impact is minimal, our goal is to ensure that these representations do not degrade the estimates further.

\paragraph{Extended experimental setup:} 
The experimental setup remains largely consistent with the previous experiments, with a few notable exceptions. We now evaluate the learners across four settings: (1) with perfect knowledge of the text-based confounders, (2) using pre-trained BioLord embeddings, (3) using pre-trained MPNet embeddings, and (4) with no access to the confounders. Both BioLord \cite{biolord} and MPNet \cite{MPNet} are sentence transformer models, from which text-based confounder representations ($\Phi_{text}$) were obtained by encoding each sentence in the clinical text notes and applying mean pooling (panel (a) of Figure~\ref{fig:models}). BioLord is a domain-specific encoder finetuned on biomedical texts and is based on MPNet, a sentence embedding model trained on extensive sentence-level datasets using a self-supervised contrastive learning objective. Based on our previous results, we focus on training sets of 300 and 3,000 samples, representing conditions where access to the text-based confounders had minimal and substantial impact on the CATE estimates of our meta-learners, respectively (see Figure~\ref{fig:baselines} in Appendix~\ref{appendix:results}).

\paragraph{Results and discussion:}

\begin{figure}[]
    \centering
    \makebox[\textwidth][c]{\includegraphics[width=1.2\textwidth]{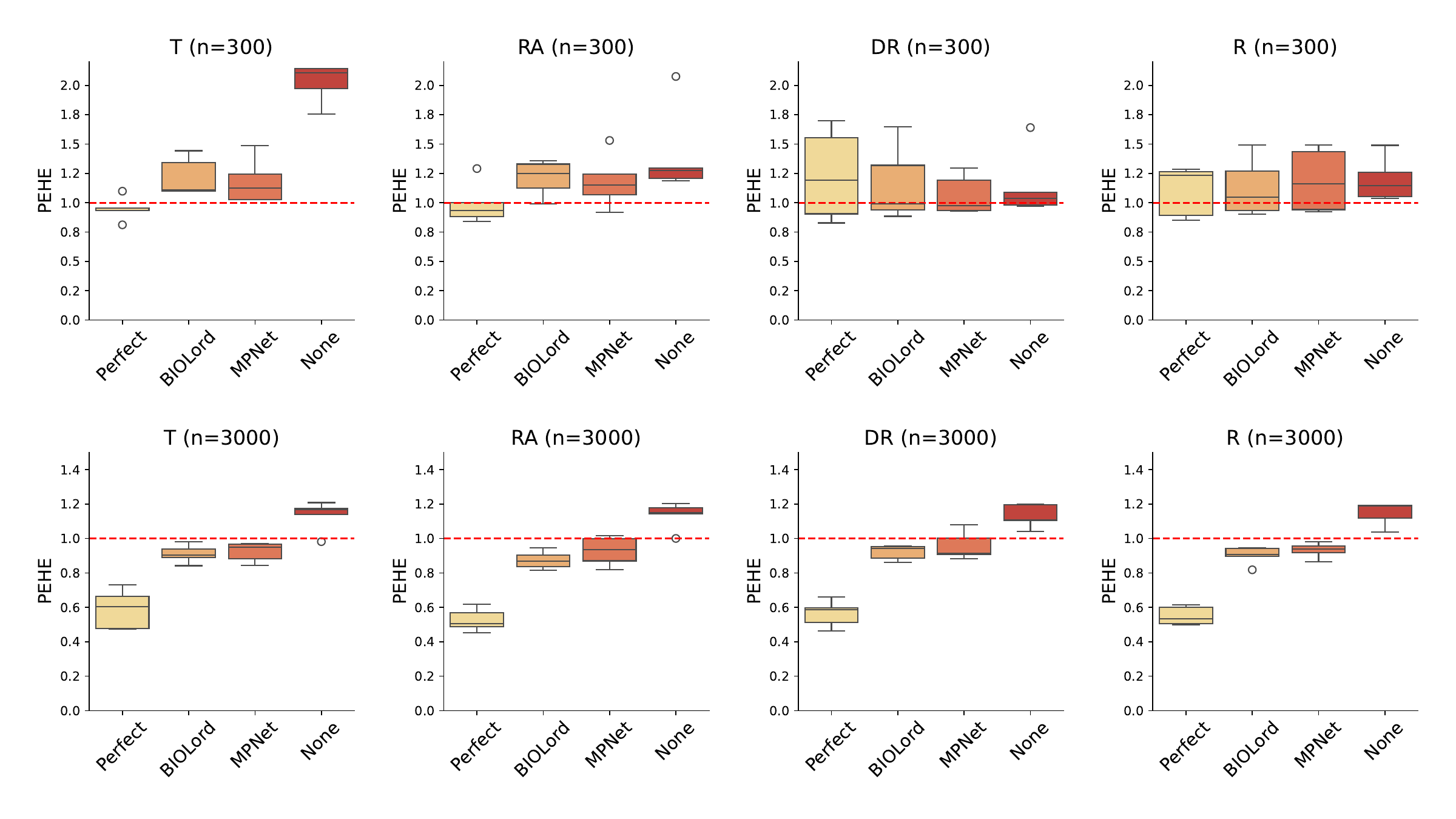}}
    \vspace{-0.9cm}
    \caption{\label{fig:embeddings} Performance comparison of the meta-learners across four settings: the text-based confounders represented (1) with perfect knowledge of them, (2) as pre-trained BioLord embeddings, (3) as pre-trained MPNet embeddings and (4) with no knowledge of them. The figure shows the PEHE on the test set (lower values indicate better performance) for each learner (columns) across two training set sizes (rows). A dashed red line at PEHE = 1 is included to aid comparison across different y-axis scales. The boxplots display results over different random seeds to illustrate variability due to weight initialisation and data sampling.}
    \label{fig:results}
\end{figure}

The results are presented in Figure~\ref{fig:embeddings}
, showing the performance of each meta-learner across the different settings and training set sizes. First, we observe that meta-learners using text representations of the confounders never perform worse than those without access to the confounders (relying only on the tabular background variables). This suggests that the pre-trained embeddings, whether from BioLord or MPNet, do not degrade the performance of the learners, even in the low-data regime. Second, when the confounding information has substantial impact on the CATE estimates (i.e., with a training set of 3,000 samples), we see that the learners using text representations achieve a performance that falls between those with perfect and no knowledge of the confounders. This indicates that the pre-trained embeddings can bridge the gap, to some extent, by capturing useful confounding information for CATE estimation. Third, we observe little difference between the domain-specific BioLord embeddings and the more general MPNet embeddings, suggesting that both are equally effective at capturing the confounding information. 

Based on these observations, we hypothesise that the reason learners using text embeddings still fall short of those with perfect confounder knowledge is the entangled nature of the text representations. We assume that the embeddings capture most, if not all, of the confounding information.\footnote{This is supported by the supervised classification experiments on the SynSUM benchmark in \cite{SynSUM}.} However, this information is likely distributed across multiple dimensions, which does not align with the data-generating process in our synthetic setup that relies on disentangled, yet correlated, confounders. This points to the broader concept of causal representation learning, which seeks to uncover high-level causal variables from low-level observations \cite{scholkopf2021toward}. One potential solution to this issue would be to disentangle the confounders through additional supervision, such as training a specialised layer on top of the embeddings using labelled confounder data, but we leave this for future work.



\section{Conclusion and future work}

Our study demonstrates the potential and the limitations of pre-trained text representations in the estimation of conditional average treatment effects (CATE) when confounders are expressed in text. Meta-learners leveraging text embeddings of confounders—whether from domain-specific or general-purpose encoders—outperform those without access to confounders (relying solely on tabular background variables). However, they still fall short of models with perfect confounder knowledge, likely due to the entangled nature of the text representations.

A first direction for future work involves addressing this entanglement. This could be achieved, for instance, through supervision by incorporating labelled data on the true confounders, by exploring causal fine-tuning strategies for text encoders (in line with the work of Veitch et al. \cite{veitch2020adapting}), or by applying techniques from causal representation learning \cite{scholkopf2021toward}.

Second, while our current work is primarily empirical, we aim to formalise our findings by investigating the role of representation error in confounders from a theoretical perspective. Specifically, we aim to study how entangled representations ($\Phi_{text}$) of text-based confounders affect CATE estimates of different learners. This exploration will shift the focus from traditional theoretical work on estimation errors due to finite samples (e.g., see \cite{curth2021really, kennedy2023towards, nie2021quasi}) to estimation errors arising from imperfect confounder representations, aligning with the work of Melnychuk et al. \cite{melnychuk2023bounds}.

Finally, a more immediate direction of future work is to explore how meta-learners perform when confounders are expressed in other modalities, such as images. In the context of the SynSUM benchmark \cite{SynSUM}, where confounders are currently expressed in clinical text notes, we could for instance augment this dataset by expressing the confounders in synthetic medical images. This could offer additional insights into the applicability of meta-learners in practical contexts.  

\section*{Acknowledgments}
Henri Arno and Paloma Rabaey’s research is funded by the Research Foundation Flanders (FWO Vlaanderen) with grant numbers 11Q2C24N and 1170124N. This research also received funding from the Flemish government under the “Onderzoeksprogramma Artificiele Intelligentie (AI) Vlaanderen” program.

\bibliographystyle{plain}
\bibliography{bibliography}

\begin{thebibliography}{10}

\bibitem{alaa2024conformal}
Ahmed~M. Alaa, Zaid Ahmad, and Mark van~der Laan.
\newblock Conformal meta-learners for predictive inference of individual treatment effects.
\newblock In {\em Advances in Neural Information Processing Systems}, 2023.

\bibitem{gausian_processes}
Ahmed~M. Alaa and Mihaela van~der Schaar.
\newblock Bayesian inference of individualized treatment effects using multi-task gaussian processes.
\newblock In {\em Advances in Neural Information Processing Systems}, 2017.

\bibitem{athey}
Susan Athey, Guido~W. Imbens, Jonas Metzger, and Evan Munro.
\newblock Using wasserstein generative adversarial networks for the design of monte carlo simulations.
\newblock {\em Journal of Econometrics}, 240(2), 2024.

\bibitem{curth2021really}
Alicia Curth, David Svensson, Jim Weatherall, and Mihaela van~der Schaar.
\newblock Really doing great at estimating cate? a critical look at ml benchmarking practices in treatment effect estimation.
\newblock In {\em Proceedings of the Neural Information Processing Systems Track on Datasets and Benchmarks}, 2021.

\bibitem{curth2021nonparametric}
Alicia Curth and Mihaela van~der Schaar.
\newblock Nonparametric estimation of heterogeneous treatment effects: From theory to learning algorithms.
\newblock In {\em Proceedings of The 24th International Conference on Artificial Intelligence and Statistics}, 2021.

\bibitem{curth2021inductive}
Alicia Curth and Mihaela van~der Schaar.
\newblock On inductive biases for heterogeneous treatment effect estimation.
\newblock In {\em Advances in Neural Information Processing Systems}, 2021.

\bibitem{fisherinverse}
Aaron Fisher.
\newblock Inverse-variance weighting for estimation of heterogeneous treatment effects.
\newblock In {\em Proceedings of The 41st International Conference on Machine Learning}, 2024.

\bibitem{frauen2024}
Dennis Frauen, Konstantin Hess, and Stefan Feuerriegel.
\newblock Model-agnostic meta-learners for estimating heterogeneous treatment effects over time, 2024.
\newblock arXiv:2407.05287v1 preprint.

\bibitem{hill2011bayesian}
Jennifer~L. Hill.
\newblock Bayesian nonparametric modeling for causal inference.
\newblock {\em Journal of Computational and Graphical Statistics}, 20(1), 2011.

\bibitem{jonkers2024conformal}
Jef Jonkers, Jarne Verhaeghe, Glenn~Van Wallendael, Luc Duchateau, and Sofie~Van Hoecke.
\newblock Conformal convolution and monte carlo meta-learners for predictive inference of individual treatment effects, 2024.
\newblock arXiv:2402.04906v4 preprint.

\bibitem{kennedy2023towards}
Edward~H Kennedy.
\newblock Towards optimal doubly robust estimation of heterogeneous causal effects.
\newblock {\em Electronic Journal of Statistics}, 17(2), 2023.

\bibitem{kingma2014adam}
Diederik~P. Kingma and Jimmy Ba.
\newblock Adam: {A} method for stochastic optimization.
\newblock In {\em Proceedings of The 3rd International Conference on Learning Representations}, 2015.

\bibitem{kunzel2019metalearners}
Sören~R. Künzel, Jasjeet~S. Sekhon, Peter~J. Bickel, and Bin Yu.
\newblock Metalearners for estimating heterogeneous treatment effects using machine learning.
\newblock {\em Proceedings of the National Academy of Sciences}, 116(10), 2019.

\bibitem{melnychuk2023bounds}
Valentyn Melnychuk, Dennis Frauen, and Stefan Feuerriegel.
\newblock Bounds on representation-induced confounding bias for treatment effect estimation.
\newblock In {\em Proceedings of The 12th International Conference on Learning Representations}, 2024.

\bibitem{morzywolek2023general}
Pawel Morzywolek, Johan Decruyenaere, and Stijn Vansteelandt.
\newblock On weighted orthogonal learners for heterogeneous treatment effects, 2024.
\newblock arXiv:2303.12687v2 preprint.

\bibitem{nie2021quasi}
Xinkun Nie and Stefan Wager.
\newblock {Quasi-oracle estimation of heterogeneous treatment effects}.
\newblock {\em Biometrika}, 108(2), 2020.

\bibitem{SynSUM}
Paloma Rabaey, Henri Arno, Stefan Heytens, and Thomas Demeester.
\newblock Synsum -- synthetic benchmark with structured and unstructured medical records, 2024.
\newblock arXiv:2409.08936v1 preprint.

\bibitem{biolord}
Fran{\c{c}}ois Remy, Kris Demuynck, and Thomas Demeester.
\newblock {B}io{LORD}: Learning ontological representations from definitions for biomedical concepts and their textual descriptions.
\newblock In {\em Findings of the Association for Computational Linguistics: EMNLP}, 2022.

\bibitem{rubinneyman}
Donald~B. Rubin.
\newblock Causal inference using potential outcomes.
\newblock {\em Journal of the American Statistical Association}, 100(469), 2005.

\bibitem{scholkopf2021toward}
Bernhard Schölkopf, Francesco Locatello, Stefan Bauer, Nan~Rosemary Ke, Nal Kalchbrenner, Anirudh Goyal, and Yoshua Bengio.
\newblock Toward causal representation learning.
\newblock {\em Proceedings of the IEEE}, 109(5), 2021.

\bibitem{shalit3}
Uri Shalit, Fredrik~D. Johansson, and David Sontag.
\newblock Estimating individual treatment effect: generalization bounds and algorithms.
\newblock In {\em Proceedings of the 34th International Conference on Machine Learning}, 2017.

\bibitem{shi2019adapting}
Claudia Shi, David Blei, and Victor Veitch.
\newblock Adapting neural networks for the estimation of treatment effects.
\newblock In {\em Advances in Neural Information Processing Systems}, 2019.

\bibitem{MPNet}
Kaitao Song, Xu~Tan, Tao Qin, Jianfeng Lu, and Tie-Yan Liu.
\newblock Mpnet: Masked and permuted pre-training for language understanding.
\newblock In {\em Advances in Neural Information Processing Systems}, 2020.

\bibitem{veitch2020adapting}
Victor Veitch, Dhanya Sridhar, and David Blei.
\newblock Adapting text embeddings for causal inference.
\newblock In {\em Proceedings of the 36th Conference on Uncertainty in Artificial Intelligence}, 2020.

\bibitem{wager2018estimation}
Stefan Wager and Susan Athey.
\newblock Estimation and inference of heterogeneous treatment effects using random forests.
\newblock {\em Journal of the American Statistical Association}, 113(523), 2018.

\bibitem{yoon2018ganite}
Jinsung Yoon, James Jordon, and Mihaela van~der Schaar.
\newblock {GANITE}: Estimation of individualized treatment effects using generative adversarial nets.
\newblock In {\em Proceedings of The 6th International Conference on Learning Representations}, 2018.

\end{thebibliography}

\appendix
\section*{Appendix}

\section{Pseudo-outcomes for the considered meta-learners}
\label{appendix:POs}

In this appendix, we provide additional details on the pseudo-outcomes of the meta-learners considered in our study. Specifically, we present the formal expressions for the pseudo-outcomes associated with the RA-learner and DR-learner, and we discuss how the R-learner can be formulated as a weighted pseudo-outcome regression. Recall that these pseudo-outcomes rely on the estimated nuisance parameters $\hat{\eta}(X) = \{\hat{\mu}_0(X), \hat{\mu}_1(X), \hat{\mu}(X), \hat{\pi}(X)\}$, obtained in the first step of the meta-learning process.

The pseudo-outcome for data instance $\left(X_i, T_i, Y_i^{obs}\right)$ according to the RA-learner is given by:
\begin{equation} 
\tilde{\tau}_{RA, i} = T_i (Y_i^{obs} - \hat{\mu}_0(X_i)) + (1 - T_i) (\hat{\mu}_1(X_i) - Y_i^{obs}) 
\end{equation}

and the corresponding pseudo-outcome for the DR-learner is given by:

\begin{equation} 
\tilde{\tau}_{DR, i} = \left( \frac{T_i}{\hat{\pi}(X_i)} - \frac{1 - T_i}{1 - \hat{\pi}(X_i)} \right) Y_i^{obs} + \left(1 - \frac{T_i}{\hat{\pi}(X_i)} \right) \hat{\mu}_1(X_i) - \left(1 - \frac{1 - T_i}{1 - \hat{\pi}(X_i)} \right) \hat{\mu}_0(X_i)
\end{equation}

Whenever we have correctly specified nuisance parameters, these pseudo-outcomes are unbiased approximations of the CATE meaning that $\mathbb{E}[\tilde{\tau}|X] = \tau(X)$. Specifically, the RA-learner’s pseudo-outcome is unbiased for the CATE given accurate estimates of the conditional outcome nuisance parameters (i.e. when $\hat{\mu}_t(X)=\mathbb{E}[Y|T=t, X]$). The DR-learner is doubly robust meaning that its potential outcome remains unbiased if either the conditional outcome nuisance parameters or the propensity nuisance parameter are correctly specified (i.e. when $\hat{\mu}_t(X)=\mathbb{E}[Y|T=t, X]$ or when $\hat{\pi}(X)=P(T=1|X)$). For a theoretical analysis of these learners, we refer the reader to \cite{curth2021nonparametric}.

\newpage
The R-learner estimates the CATE by minimising the following loss function:

\begin{equation}
    \arg \min_{\hat{\tau}} \sum_{i=1}^{D} \left[ \left(Y_i^{obs} - \hat{\mu}(X_i)\right) - \left(T_i - \hat{\pi}(X_i)\right) \hat{\tau}(X_i)\right]^2 
\end{equation}

Alternatively, the R-learner can also be formulated as fitting a weighted regression on a pseudo-outcome defined as:
\begin{equation}
    \tilde{\tau}_{R, i} = \frac{Y_i^{obs} - \hat{\mu}(X_i)}{T_i - \hat{\pi}(X_i)}
\end{equation}
with weights $(T_i - \hat{\pi}(X_i))^2$ and the squared error loss function \cite{fisherinverse}. For the theoretical background on the R-learner, we refer the reader to \cite{nie2021quasi}. Note that we use the latter approach in our experiments.

The learners considered in our study are among the most prominent for CATE estimation but have been developed from very different perspectives. Recently, Morzywołek et al. studied the connection between these learners and proposed a single unifying framework. For more details, we refer the reader to \cite{morzywolek2023general}.

\section{Example entry from the SynSUM dataset}
\label{app:SynSum}

\begin{figure}[h]
    \centering
    \makebox[\textwidth][c]{\includegraphics[width=1.13\textwidth]{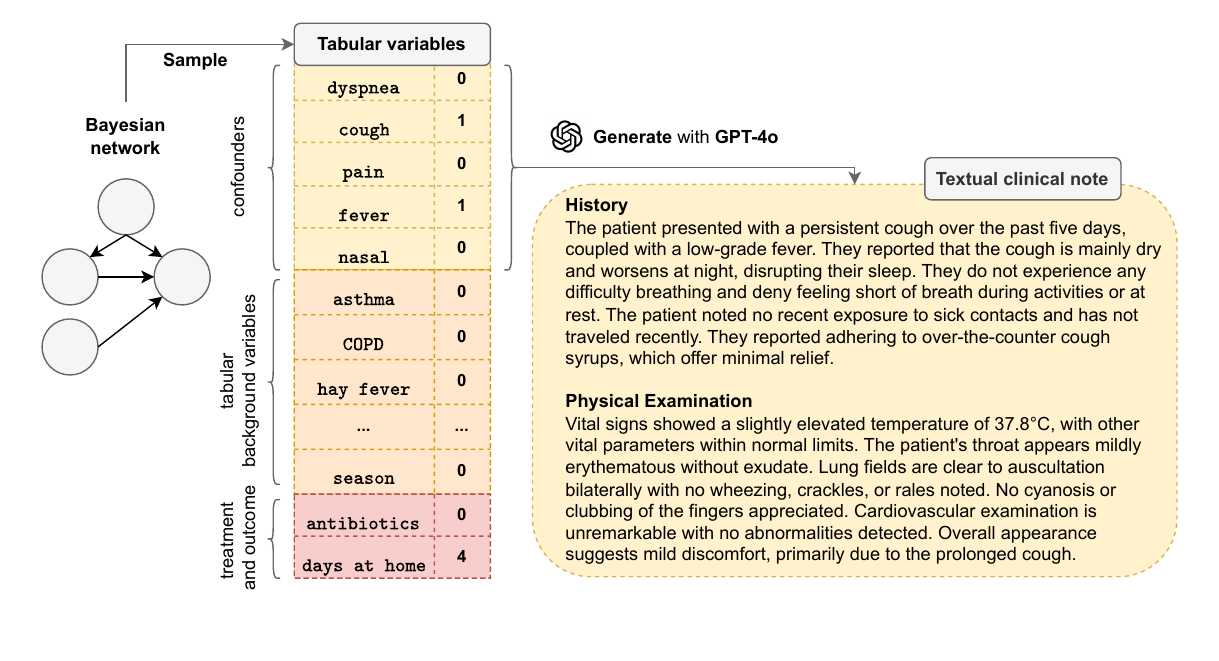}}
    \caption{\label{fig:synsum} An example entry from the SynSUM dataset that combines structured tabular variables, sampled from a Bayesian network, with a textual clinical note, generated by GPT-4o. This synthetic example represents a realistic patient record from a patient encounter in primary care containing both structured data (e.g., underlying conditions of the patient) and unstructured text (e.g. describing the results of a physical examination revealing the symptoms a patient experiences).}
\end{figure}

In Figure~\ref{fig:synsum} of this appendix, we present an example entry from the SynSUM dataset \cite{SynSUM}, containing structured tabular variables sampled from a Bayesian network and a textual clinical note generated by GPT-4o. This illustrates a synthetic patient record composed of both structured and unstructured data.

\newpage
\section{Training details of the nuisance parameter models and the second-stage regressors}
\label{app:training_details}

The nuisance parameter models consist of four separate multi-layer perceptrons (MLPs), each dedicated to estimating a specific nuisance parameter with a distinct target and loss function. For estimating $\hat{\mu}(X)$,  we use the mean squared error loss with the observed outcome $Y^{obs}$ as target. For $\hat{\mu}_t(X)$, we also compute the mean squared error, but only using samples from the respective treatment groups ($T=t$) with the observed outcome $Y^{obs}$ as target. Finally, for $\hat{\pi}(X)$, we use binary cross-entropy loss with the treatment indicator $T$ as target.

Each MLP is trained with its own Adam optimizer \cite{kingma2014adam} and learning rate scheduler. The learning rate scheduler reduces the initial learning rate by a factor of 0.1 if the validation loss does not improve for 5 consecutive epochs. A randomly sampled 20\% of the training set serves as the validation set. The training lasts for 75 epochs with batches of 32 samples, during which the model alternates between tasks for each nuisance parameter in every batch. L2-regularization with a weight decay of 1e-4 is applied. The initial learning rate for each head was tuned separately as a hyperparameter (the initial learning rate that minimised the validation loss was ultimately selected). This training procedure is consistently applied across all settings discussed in the paper.

The second-stage regressors are trained similarly, with the primary difference being the use of pseudo-outcomes $\tilde{\tau}$ as targets (and mean squared error as the loss function). Unlike the nuisance parameter models, the second-stage regressor does not alternate tasks between batches. The entire training process—which includes the nuisance parameter model and second-stage regressor—is repeated five times for each learner and training set size, with different random seeds to account for variations in weight initialisation and data sampling.

\section{Experimental results}
\label{appendix:results}

This appendix presents the results from our initial experiments, showcasing the performance of our considered meta-learners under different settings and across various training set sizes. The settings correspond to different representations of the text-based confounders, 
(1) with perfect knowledge and (2) with no knowledge of them. 

\begin{figure}[]
    \centering
    \makebox[\textwidth][c]{\includegraphics[width=1.13\textwidth]{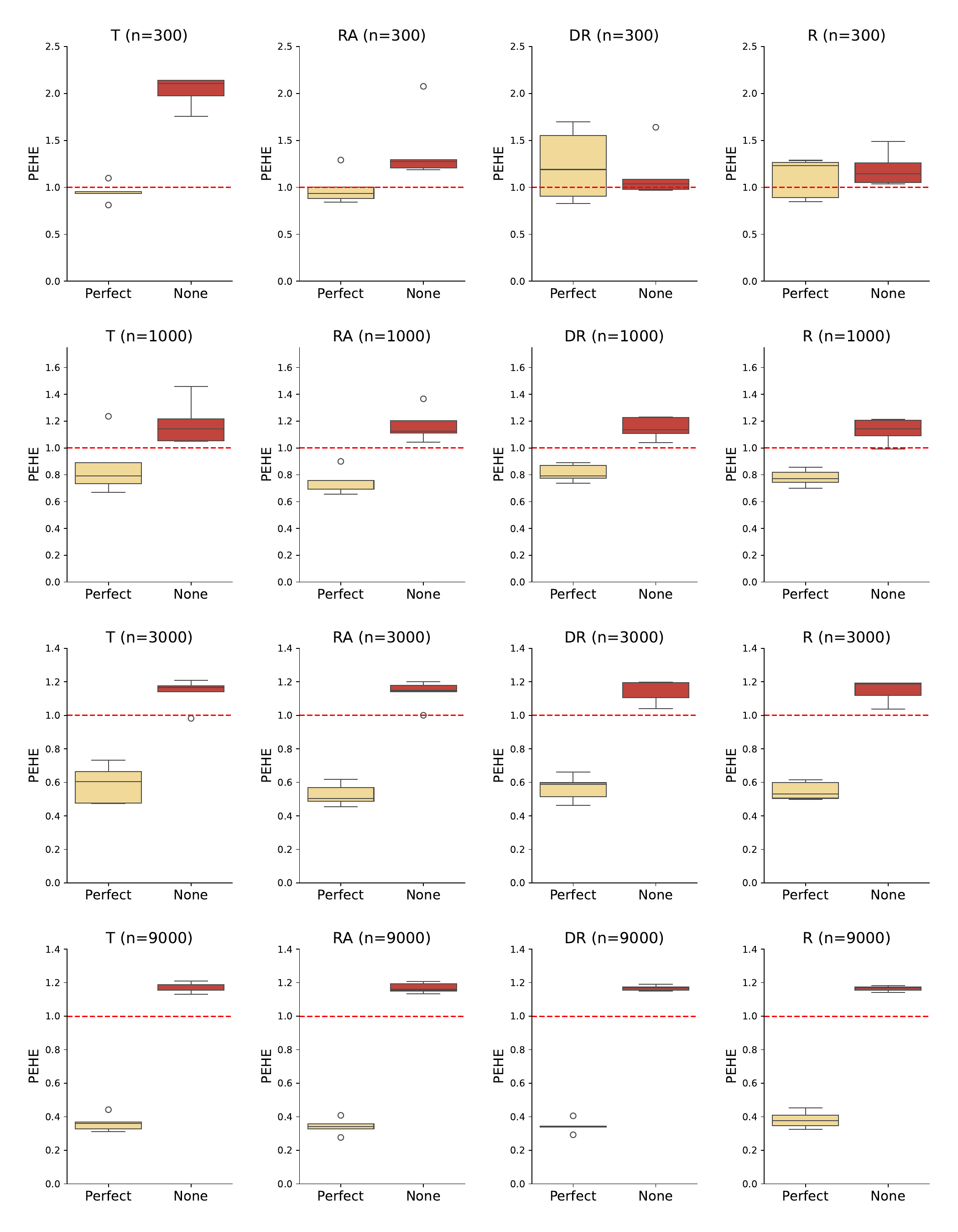}}
    \caption{\label{fig:baselines} Performance comparison of the meta-learners across two settings: the text-based confounders represented (1) with perfect knowledge of them and (2) with no knowledge of them. The figure shows the PEHE on the test set (lower values indicate better performance) for each learner (columns) across four training set sizes (rows). A dashed red line at PEHE = 1 is included to aid comparison across different y-axis scales. The boxplots display results over different random seeds to illustrate variability due to weight initialisation and data sampling.}
\end{figure}


\end{document}